\documentclass{article}
\usepackage{arxiv}
\usepackage[T1]{fontenc}
\usepackage[utf8]{inputenc}
\usepackage{amsmath,amssymb,booktabs,array,graphicx,microtype}
\usepackage[numbers,sort&compress]{natbib}
\usepackage{xurl}
\usepackage[unicode,colorlinks=true,citecolor=blue,linkcolor=blue,urlcolor=blue]{hyperref}
\AtBeginDocument{\raggedbottom\setlength{\headheight}{16pt}}
\renewcommand{\headeright}{}
\renewcommand{\undertitle}{}
\renewcommand{\shorttitle}{Emergence of Action Spaces}
\title{The Emergence of Action Spaces in Reinforcement Learning}
\author{Dengzhe Hou\textsuperscript{1,2}, Fangzhou Lin\textsuperscript{1,3},
  Xueying Zhan\textsuperscript{2}, and
  Kazunori D Yamada\textsuperscript{1,2*}\\[0.4em]
  {\small\textsuperscript{1}Unprecedented-scale Data Analytics Center, Tohoku University, Sendai, 980-8572, Japan}\\
  {\small\textsuperscript{2}Graduate School of Information Sciences, Tohoku University, Sendai, 980-8579, Japan}\\
  {\small\textsuperscript{3}Department of Computer Science and Engineering, Texas A\&M University, College Station, TX 77843, USA}\\
  {\small\textsuperscript{*}\texttt{yamada@tohoku.ac.jp}}}
\date{}
\newcommand{\YamCartPole}{CartPole}
\newcommand{\YamDoorKey}{DoorKey}
\newcommand{\YamForceGrid}{Gridworld}
\newcommand{\YamNumericRows}{%
\YamDoorKey  & 3 & 1,009,809 & 568,909 & 901 & 12/20\\
             & 4 & 1,551,611 & 1,183,403 & 944 & 13/20\\
             & 5 & 1,592,637 & 1,179,103 & 1,447 & 11/20\\
\midrule
\YamForceGrid& 3 & 561,366 & 4,545 & 26 & 18/20\\
             & 4 & 596,786 & 7,545 & 32 & 13/20\\
             & 5 & 593,488 & 4,748 & 26 & 19/20\\
\midrule
\YamCartPole & 3 & 2,428,165 & 1,507,125 & 383 & 6/20\\
             & 4 & 2,392,054 & 1,384,205 & 963 & 14/20\\
             & 5 & 2,587,046 & 1,398,890 & 1,024 & 9/20\\
}

\begin{document}
\maketitle

\begin{abstract}
In reinforcement learning, the form of the actions an agent can produce is usually specified before training. In this study, we developed YamRL, a general-purpose method for generating action spaces that learns through interaction with the environment which values to produce as actions. A recurrent neural network generates byte sequences, and a shared decoder converts them into values. The environment determines whether it can execute each value. We used the same YamRL configuration and the same reinforcement learning algorithm in three tasks that required an integer command, a two-component control signal, and a floating-point force, respectively. We trained on each task separately and solved all three tasks. These results show that a common initial mechanism can acquire effective actions of different types and dimensions through interaction with the environment.
\end{abstract}


\section{Introduction}

In reinforcement learning, the specifications of the action interface usually determine which candidates an agent can try. An agent in a cart-pole task may be given two commands or asked to specify a scalar force. An agent in a gridworld may be given a list of movement directions. \underline{Y}amada \underline{m}odified \underline{R}einforcement \underline{L}earning (YamRL), developed in this study, uses a common set of bytes to build candidates. A shared decoder allows the generated byte sequences to represent individual values such as integers, floating-point numbers, strings, and Boolean values, as well as structures such as lists, dictionaries, and matrices. The environment defines the observations, valid inputs, state transitions, and rewards. The policy receives no action list, action type, dimensionality, range, or environment identifier. We investigated whether this common mechanism could learn to generate valid values in environments with different input requirements.

Markov decision processes with unawareness model how useful actions are discovered through an exploration action. In a robotics application, gaits are acquired after specifying joint coordinates, ranges of motion, and discretization \cite{halpern2010,rong2016}. Action embeddings learn the structure of an existing action set, and symbolic policy search generates control expressions for predefined action coordinates \cite{chandak2019,landajuela2021}. CALM generates text commands using a language model trained on records of human gameplay \cite{yao2020}. Gato mentions raw byte sequences as a possible representation. In its experiments, Gato tokenized data according to its type and used the environment's action specification to determine the output length \cite{reed2022}. YamRL generates actions using the same set of bytes and the same rule for delimiting sequences in every environment and learns the policy through interaction with each environment.

\section{Methods}
\subsection{Interaction and action acquisition}

In YamRL, the network that generates action candidates from observations serves as the policy. Its parameters are learned using Advantage-Weighted Regression (AWR) \cite{peng2019}.

Let \(t\) index candidate generation, \(o_t\) denote the observation at that point, and \(\theta\) denote the trainable parameters of the policy. YamRL takes the current observation \(o_t\) and generates a byte sequence \(b_t\) according to the policy \(\pi_\theta\). The shared decoder \(D\) converts the sequence to one typed value \(x_t=D(b_t)\) or rejects it. Each environment determines whether it can execute the decoded value according to its input requirements. When the environment accepts a value, it advances one step and returns an observation, a scalar reward, and information on whether the episode has ended. Incomplete byte sequences, invalid syntax, and values that fail to meet the input requirements are rejected. When a candidate is rejected, the environment stays in the same state and the shared learning algorithm assigns a reward of \(-0.1\). The candidate index \(t\) also advances after a rejection.

Learning changes the probability of generating each candidate given the observation. We use the term action-space acquisition for learning to generate operations that the environment can execute and that help solve the task. After training, YamRL uses the same procedure to generate a candidate whenever it receives an observation.

\subsection{Byte-sequence generation and decoding}

The shared observation encoder uses signed hashing to map an observation to a 96-component vector. Field names and indices in the observation distinguish the component positions. Each finite numerical value \(z\) in the observation is transformed to \(z/(1+|z|)\), and each string is hashed in its entirety. A trainable affine transformation and \(\tanh\) are used to form a 32-dimensional hidden state, which is initialized for every candidate. The trainable start vector and the embedding for each byte are 16-dimensional. YamRNN updates the hidden state each time it receives an input, and an affine transformation gives log weights for the 256 possible bytes \cite{yamada2021}. The recurrent cell has 2,112 trainable parameters, and the full model has 17,809, including the observation transformation, embeddings, byte output layer, and value function.

Each byte is drawn from the categorical distribution produced by the network. The byte output layer is initialized to zero, so the conditional distribution is uniform over the 256 possible bytes before training. Candidate generation stops when one value can be decoded, when a syntax error is detected, or when the sequence reaches 64 bytes. Let \(o\) be the observation, \(L\) the number of bytes in the candidate, and \(b_j\) the \(j\)th byte. The candidate is written as \(b=(b_1,\dots,b_L)\). Here \(b_{<j}\) is the sequence preceding byte \(j\), and \(p_\theta(b_j\mid o,b_{<j})\) is the probability of selecting \(b_j\) given the observation and preceding bytes. The log probability of the candidate used for learning is computed as
\[
\log\pi_\theta(b\mid o)
=\sum_{j=1}^{L}\log p_\theta(b_j\mid o,b_{<j}).
\]
The type and length of the value are determined without consulting an environment-specific set of candidates.

The shared decoder is based on MessagePack \cite{msgpack}. It handles scalars, structured values such as lists, and strings. The same modification to the assignment of integer codes is used for every task. Integers are ranked by the smallest width that can represent them in standard MessagePack, and the ranks are mapped to \(0,-1,1,-2,2,\ldots\) using the ZigZag rule. The upper end of the 64-bit integer range is retained. The 160 one-byte integer codes correspond to \([-80,79]\), giving positive and negative ranges of nearly equal size. The same assignment is used for integers in lists. Floating-point numbers are read using the MessagePack IEEE 754 representation.

\subsection{Learning}

Observations, generated byte sequences, and rewards are stored as experiences. At the end of each episode, the Monte Carlo return \(G_t\) is computed using a per-candidate discount factor of \(\gamma=0.99\), including rejected candidates. Up to 50,000 experiences are retained. Every 32 candidate generations, if at least 32 experiences are available, 32 are sampled uniformly for an update.

Let \(V\) denote the value function. First, the mean squared error of the value function is reduced using \(G_t\) as the target. Next, \(G_t-V(o_t)\) is computed using the updated value function and standardized using the minibatch mean and population standard deviation. A value of \(10^{-5}\) is added to the denominator. The temperature parameter \(\beta\) is set to \(0.05\) for all three tasks, as in the original AWR paper \cite{peng2019}. Let \(\widehat A_t\) be the standardized value and \(U\) the set of indices for the 32 experiences sampled in the minibatch. The following loss is minimized to update the policy:
\[
F(\theta)=-\frac1{32}\sum_{t\in U}
\min\!\left(20,\exp(\widehat A_t/\beta)\right)\log\pi_\theta(b_t\mid o_t).
\]
Both updates use Adam with a learning rate of \(10^{-3}\) \cite{kingma2015}. The gradient norm is clipped at 1. The update rules, network, and initialization procedure are shared across the three tasks.

\section{Experiments}
\subsection{Three tasks with different input requirements}

We used the existing DoorKey environment \cite{minigrid}. We passed the observations returned by the environment to the common observation encoder. The validity of integer commands was checked on the environment side, and accepted commands were executed in the original environment. An episode was successful if the agent picked up the key, unlocked the door, and reached the goal within 250 accepted steps. For an accepted action, \(s\) and \(s'\) denote the states before and after execution, respectively. We denote the progress score assigned to state \(s\) by \(\Phi(s)\). The score was \(0\) before unlocking without the key, \(0.25\) before unlocking while holding the key, and \(0.5\) after unlocking. We added \(0.99\Phi(s')-\Phi(s)\) to the original reward and set \(\Phi(s')=0\) when the goal was reached \cite{ng1999}.

We built Gridworld as an \(8\times8\) grid. It returned the coordinates of the current position as its observation and accepted a list of two finite numbers as a control signal. We applied \(\tanh\) to each component, rotated the resulting vector by a fixed angle, and added it to the current position. We then rounded the coordinates and clipped them to the grid boundaries. The agent started in one of the four upper-left cells. An episode was successful if the agent reached the goal in the diagonally opposite corner within 128 accepted steps. We defined the state \(s\) as the current grid coordinates and \(g\) as the goal coordinates, and set the progress score to \(\Phi(s)=-0.25\lVert s-g\rVert_2/(7\sqrt{2})\). In addition to a goal reward of 1, the environment gave a progress-dependent reward at each accepted step using the same expression as in DoorKey. The rotation angle, the dimensionality of the control signal, and the goal coordinates were used only by the environment.

We used the equations of motion for CartPole implemented in Gymnasium 1.4.0 \cite{gymnasium_cartpole}. We let the environment accept a continuous-valued force applied to the cart. The observation comprised the position and velocity of the cart and the angle and angular velocity of the pole. We drew the initial angle from a uniform distribution between \(-10^\circ\) and \(10^\circ\) and drew each of the other three components from a uniform distribution over \([-0.05,0.05]\). At each step, the environment accepted one finite floating-point number in the range \([-10^6,10^6]\) N, including 0 N. Each accepted step advanced the simulation by 0.02 seconds and yielded a reward of \(+1\). An episode ended in failure if the absolute cart position exceeded 2.4 m or the absolute pole angle exceeded \(12^\circ\). Maintaining control for 500 accepted steps counted as success.

\subsection{Training and frozen evaluation}

For each task, we trained policies independently with seeds 3, 4, and 5. We limited the number of candidates generated in each episode to 100,000.

Training stopped once 16 of the most recent 20 episodes had succeeded. To measure the success rate of the trained policy, we froze the weights and evaluated it for 20 episodes using reset and sampling seeds distinct from those used for training. Let \(m\) denote the model seed and \(i\) the episode index starting at 0. We used \(1000m+i\) as the training reset seed. For evaluation, we used \(300000+100(m-3)+i\) for resets and \(m+1+10000i\) for sampling. During evaluation, we continued to sample candidates from the current observation.

The code used in the experiments and videos showing an example run for each of the three tasks were released on GitHub (\url{https://github.com/caramelazuki/YamRL}).

\section{Results}
\subsection{Success on the three tasks}

The agent solved every task repeatedly, and all three seeds for each of the three tasks met the stopping criterion for training. Evaluation with frozen weights after training yielded 36 successes out of 60 episodes for DoorKey, 50 out of 60 for Gridworld, and 29 out of 60 for CartPole. Each of the three seeds succeeded at least once on every task.

In Table~\ref{tab:numeric}, the candidate count is the total number of byte sequences generated during training, including sequences that could not be decoded and candidates rejected by the environment. The accepted step count is the number of times the environment executed a candidate, including cases in which the position or similar state variables remained unchanged after execution. Training successes are the total number of successful episodes during training. Frozen evaluation reports the number of successes in 20 episodes run with the trained weights fixed and with random seeds distinct from those used for training.

\begin{table}[!htbp]
\centering
\caption{Results for the three numerical tasks.}
\label{tab:numeric}
\begin{tabular*}{\linewidth}{@{\extracolsep{\fill}}llrrrr@{}}
\toprule
Environment & Seed & Candidates & Accepted steps & Training successes & Frozen evaluation\\
\midrule
\YamNumericRows
\bottomrule
\end{tabular*}
\end{table}

Across the three seeds, the acceptance rates for candidates generated during training were \(70.6\%\) for DoorKey, approximately \(1.0\%\) for Gridworld, and \(57.9\%\) for CartPole.

Gridworld was solved repeatedly with an acceptance rate of about 1\% because the agent could try another candidate from the same position after a rejection. These results cover three tasks learned independently using a common mechanism. Open questions include whether a trained policy can transfer to a different environment and how many candidates are needed compared with training in which the action specification is provided.

\subsection{Values used in successful episodes}

We decoded the frozen evaluation records and counted the values accepted in successful episodes. DoorKey used seven distinct integer commands in 36 episodes. Gridworld accepted 3,314 lists across 50 episodes, including 2,959 distinct pairs of numbers. CartPole received 14,500 force inputs comprising 14,090 distinct force values across 29 episodes. The forces ranged from approximately \(-32.00\) N to \(31.06\) N. In Gridworld, saturation of the control signal and rounding of the destination can cause different pairs of numbers to produce the same move.

In successful episodes, for example, DoorKey executed the integer 2 decoded from \texttt{04} as a forward command. Gridworld decoded \texttt{92 67 38} to \([-52,28]\) and executed it as a two-component control signal. CartPole applied a force of approximately \(-1.792\ {\rm N}\), decoded from \texttt{ca bf e5 6c 7e}, to the cart. Gridworld rejected the single integer \(-45\), and CartPole rejected a force of approximately \(-6.00\times10^{12}\ {\rm N}\).

\section{Discussion}

To the best of our knowledge, this study is the world's first empirical demonstration that an agent using a common byte-sequence generator and decoder can acquire effective operations in different formats through interaction with the environment and solve multiple tasks without supplying the policy with action types, dimensions, or ranges in advance. The same policy output layer and decoder can be used in environments that require different formats, and learning can change the byte sequences that the policy generates.

The shared decoder can represent integers, floating-point numbers, strings, and Boolean values as individual values or as elements of lists or dictionaries (hash maps). Matrices can be represented as nested lists. The same generation mechanism could also be applied to environments that accept these formats as actions.

YamRL is an action generation mechanism that generates byte sequences from observations and sends the decoded values to the environment. A learning algorithm that updates the policy using experience from interaction with the environment can be chosen independently of the action generation procedure. We used AWR for the three tasks in this study, and other reinforcement learning algorithms can also be used.

The action that a generated value produces depends on the interface with the environment. Motors that move a body, device controllers, and software APIs are examples of interfaces that link inputs to changes in the environment. In environments such as Gridworld, where different pairs of numbers can produce the same movement, the diversity of acquired actions can be assessed by examining both the variety of generated values and the changes in state before and after execution.

In the current implementation, each candidate is limited to 64 bytes. Representing a long string or a large matrix as a single operation therefore requires raising this limit. Raising the limit expands the range of representable values and increases the number of byte sequences that can be explored.

\bibliographystyle{unsrtnat}
\bibliography{references}
\end{document}